\documentclass{article}

\RequirePackage{natbib}
\usepackage{graphicx} 
\usepackage[utf8]{inputenc} 
\usepackage[T1]{fontenc}    
\usepackage{hyperref}       
\usepackage{url}            
\usepackage{booktabs}       
\usepackage{amsfonts}       
\usepackage{nicefrac}       
\usepackage{microtype}      
\usepackage{listings}       
\usepackage{mathptmx}       
\usepackage{xstring,etoolbox} 

\usepackage[margin=0.5cm]{caption} 

\newcommand{\specialcellcenter}[2][c]{%
  \begin{tabular}[#1]{@{}c@{}}#2\end{tabular}}

\setcitestyle{authoryear,round,citesep={;},aysep={,},yysep={;}}

\newcommand{\fixsplit}[2]{\StrLen{#2}[\mynum]\ifnumcomp{\mynum}{<}{\numexpr(#1)+1\relax}%
  {#2}%
  {\StrSplit{#2}{#1}{\myfirststr}{\mysecondstr}\myfirststr\linebreak
  \fixsplit{#1}{\mysecondstr}}}

\renewcommand{\cite}[1]{\citep{#1}}

\usepackage{blindtext}
\usepackage{geometry}
\begin{document}

{\fontfamily{ptm}\selectfont
\title{Accurate Prediction of Ligand-Protein Interaction \\ Affinities with Fine-Tuned Small Language Models}}

\author{Ben Fauber\thanks{Correspondence to: Ben.Fauber@dell.com} \\
\normalsize{Dell Technologies}
}
\date{June 26, 2024}

\maketitle

\begin{abstract}
We describe the accurate prediction of ligand-protein interaction (LPI) affinities, also known as drug-target interactions (DTI), with instruction fine-tuned pretrained generative small language models (SLMs). We achieved accurate predictions for a range of affinity values associated with ligand-protein interactions on out-of-sample data in a zero-shot setting. Only the SMILES string of the ligand and the amino acid sequence of the protein were used as the model inputs. Our results demonstrate a clear improvement over machine learning (ML) and free-energy perturbation (FEP+) based methods in accurately predicting a range of ligand-protein interaction affinities, which can be leveraged to further accelerate drug discovery campaigns against challenging therapeutic targets.
\end{abstract}

\section{Introduction}

Significant advances have been made in the \emph{in silico} prediction of molecular and pharmacokinetic properties associated with successful drug-like molecules \cite{Leeson2021TargetbasedEO, Lombardo2017InSA}. These cheminformatics advances have laid the foundation for further enhancements in drug candidate screening, prioritization for advancement into \emph{in vivo} studies, and clinical candidate selection \cite{Maurer2021DesigningSM}. Despite these impressive improvements in molecular property predictions, a considerable challenge remains in accurately predicting the affinity/potency of a ligand-protein interaction (LPI), also known as a drug-target interaction (DTI) \cite{Yamanishi2008PredictionOD}.

Drugs convey their phenotypic effects through interactions with a variety of biological targets with varying affinities \cite{Swinney2011HowWN}. Some interactions produce desirable outcomes and phenotypes, while others can create undesired side effects and/or safety risks \cite{Waring2015AnAO}. Accurately predicting the affinities of ligand-protein interactions would enable drug discovery teams to better design and prioritize the synthesis of molecules that interact with intended protein targets, while minimizing undesired interactions with off-targets like hERG and liver enzymes, ultimately increasing the chances of preclinical success. \cite{Sadybekov2023ComputationalAS}.

\subsection{Our Contribution}

In our work we used pretrained foundational small language models (SLMs), which were generative models with millions of parameters, as starting points. These small foundational models were instruction fine-tuned on domain-specific data for a few epochs. We evaluated the performance of our instruction fine-tuned language models against ground truth data (\emph{i.e.}, out-of-sample "test" data) in a zero-shot setting within a rigorous and reproducible evaluation framework.

\begin{figure}[!ht]
\vskip 0.2in
\begin{center}
\includegraphics[width=100mm]{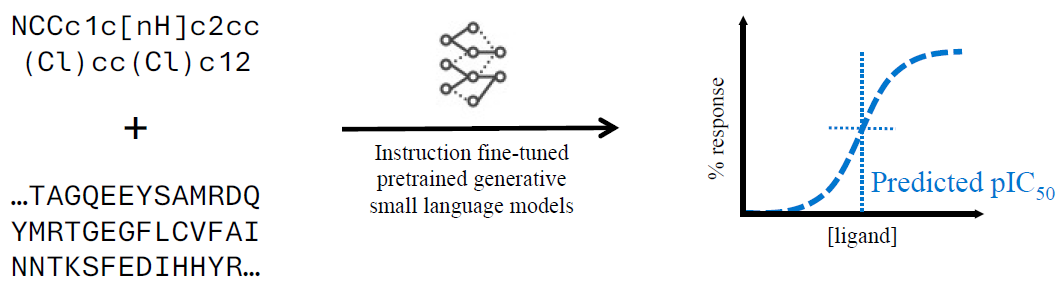}
\caption{Illustration of our proposed task: prediction of ordinal affinity values associated with ligand-protein interactions on out-of-sample data in a zero-shot setting. Only the SMILES string of the ligand (top left) and the amino acid sequence of the target protein (bottom left) are used as the model inputs.}
\label{overview}
\end{center}
\end{figure}

Herein, we demonstrate accurate prediction for a range of ordinal affinity values associated with ligand-protein interactions on out-of-sample data in a zero-shot setting. Only the SMILES (Simplified Molecular-Input Line-Entry System) string \cite{Swanson2004TheEO, Weininger1988SMILESAC} of the ligand and the amino acid sequence of the target protein were used as model inputs (Figure 1). Our results demonstrate a clear improvement over machine learning (ML) and free-energy perturbation (FEP) based methods in accurately predicting a range of ligand-protein interaction affinities, which can be further leveraged to accelerate drug discovery campaigns against challenging therapeutic targets.

\section{Related Work}

Drug discovery is a challenging multivariate optimization process \cite{Hughes2011PrinciplesOE}. Ligand-protein, or drug-target, interaction affinities are typically assessed by biochemical assays \cite{Macarron2011ImpactOH}, biophysical assays such as nuclear magnetic resonance (NMR) or surface plasmon resonance (SPR) \cite{Renaud2016BiophysicsID}, and in some instances assumed via phenotypic assays \cite{Moffat2017OpportunitiesAC}. These assay results are the gold standard by which ligands/drugs are assessed and prioritized for progression in preclinical drug discovery campaigns.

\subsection{Machine Learning and Deep Learning}

Several research groups have explored the \emph{in silico} prediction of ligand-protein interaction affinities with  statistical machine learning (ML) algorithms \cite{Oliveira2024InferringMI, Kimber2021DeepLI, doi:10.1021/acs.jcim.9b00375, Mayr2018LargescaleCO,  Martin2011ProfileQSARAN, Yamanishi2008PredictionOD, 10.1093/bioinformatics/btm580}. Many other groups have explored deep learning (DL) methods \cite{Huang2020DeepPurposeAD, Huang2020MolTransMI, Li2020MONNAM, ztrk2019WideDTAPO, Whitehead2019ImputationOA, Lee2018DeepConvDTIPO, beyondthehypeDL2017, Wen2017DeepLearningBasedDI}.

Published results have typically relied upon small data sets of approximately 10,000 or fewer examples where ligand-protein interaction affinities were represented as binary values with a "binder" represented as a $1$, and "non-binder" represented as a $0$ in the data sets. Examples of these binary ligand-protein interaction data sets include BioSNAP \cite{biosnapnets}, DrugBank \cite{Wishart2007DrugBankAK}, and the Yamanishi data set \cite{Yamanishi2008PredictionOD}.

Binary data sets and logistic regression methods offer a fruitful landscape for impressive receiver operating characteristic (ROC) curves and accuracy values, as predicting logits is a well-formulated machine learning task \cite{James2013}. Yet, in practice ligand-protein binding affinities are a continuum and not binary values. Further, binder/non-binder binary classification is of limited practical value when rank ordering virtual screening molecules for purchase and \emph{in vitro} testing, as virtual screening campaigns can include more than $10^{10}$ compounds \cite{Sadybekov2021SynthonbasedLD}.

The commonality of machine learning methods across the LPI prior art has led research groups to focus on various LPI data representations. Groups have explored vector embeddings of both the ligand \cite{Kimber2021DeepLI} and protein \cite{Kalakoti2022TransDTITL}, including dense and sparse embedding techniques. Recent studies have revealed that the data embedding method plays a minimal role in the accurate prediction of ligand-protein interaction affinities \cite{doi:10.1021/acs.jcim.3c01208}. 

Graph representations of the ligands and proteins data sets have also been explored. In this setting, nodes that meet user-defined thresholds via inner products and other similarity assessments are connected by edges \cite{Svensson2024HyperPCMRT, Chatterjee2023ImprovingTG, Thafar2022Affinity2VecDB, Kimber2021DeepLI}.  Despite these efforts, the accurate \emph{in silico} prediction and quantification of ligand-protein interactions with machine learning and/or deep learning methods remains and unsolved challenge. 

\subsection{Physics-based Methods}

Free-energy perturbation (FEP+) calculations are computationally intensive and low throughput physics-based approaches for rank ordering ligand-protein interactions \cite{Wang2015AccurateAR}. Despite these limitations, FEP+ methods are often used to supplement drug discovery campaigns as they allow practitioners to rank order LPI affinities of proposed ligands relative to a known benchmark ligand(s). 

Free-energy perturbation calculations require ligands to be bound in an X-ray cocrystal of the ligand and protein, or low-energy conformations of ligands are docked into a known X-ray structure or predicted 3D structure of the target protein (Figure 2). Predicted 3D conformations of proteins can be calculated with tools such as RoseTTAFold \cite{doi:10.1126/science.abj8754} or AlphaFold \cite{Jumper2021HighlyAP}. Ligand docking can be performed with tools such as GOLD \cite{JONES1997727} or GLIDE \cite{doi:10.1021/jm0306430}. Biological assay data is essential to correlate the \emph{in vitro} affinity of the  ligand-protein interaction with the FEP+ calculations ($\Delta G_{exp}$ in Figure 2) \cite{Ross2023TheMA}.  

\begin{figure}[!ht]
\vskip 0.2in
\begin{center}
\includegraphics[width=140mm]{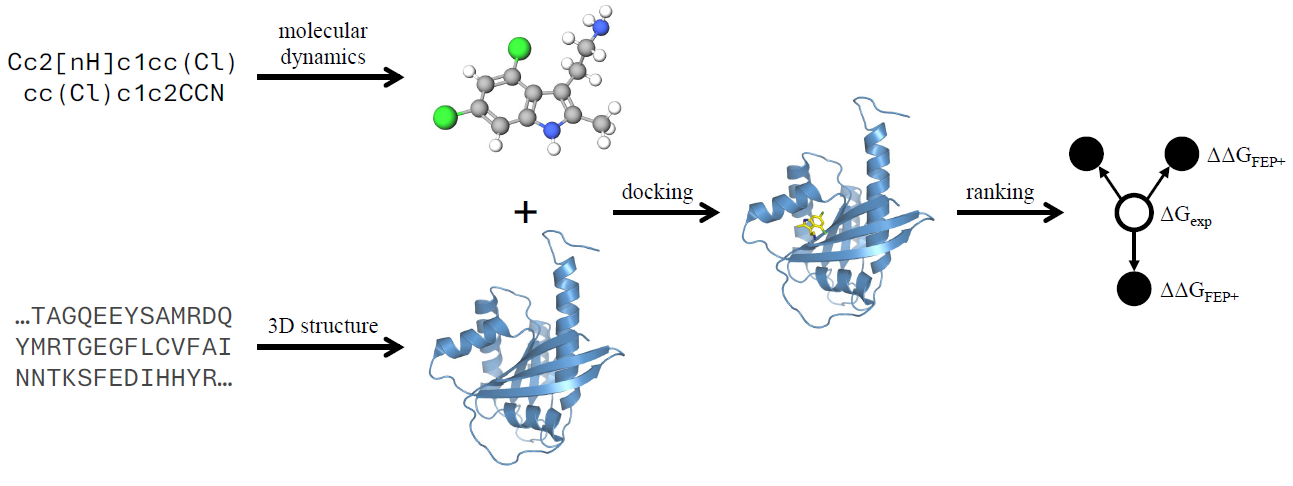}
\caption{Physics-based virtual screening to rank order ligand-protein interactions with free-energy perturbation calculations (FEP+). An example is shown with the DCAI ligand and human KRas4B-G12D protein [PDB: 4DST]. The SMILES string of the ligand must be converted into a low-energy 3D-conformation (grey). Free-energy perturbation calculations require ligands to be bound in an X-ray cocrystal of the ligand (yellow) and protein (blue), or low-energy conformations of ligands (grey) are docked into a known X-ray structure, or a predicted 3D-structure from the corresponding amino acid sequence, of the target protein (blue). Free-energy perturbation calculations often require multiple validated binders of known binding affinities to benchmark the method ($\Delta G_{exp}$) and rank order the FEP+ calculation outcomes of proposed ligands ($\Delta\Delta G_{FEP+}$) relative to the benchmark $\Delta G_{exp}$ values.}
\label{currentstate}
\end{center}
\end{figure}

Free-energy perturbation calculations often require multiple validated binders of known binding affinities to benchmark the method \cite{Ross2023TheMA}. Accurate FEP+ calculations, which can fall within a range of $\pm 1 - 3$ kcal/mol to a known benchmark depending on the protein target, followed by additional FEP+ calculations for the proposed ligands, enable the rank ordering of proposed ligands relative to the benchmark compound(s) \cite{Ross2023TheMA, doi:10.1021/acs.jcim.0c00900}. At the time of this publication, the high cost and low throughput of FEP+ methods prohibit it from being a viable method for large-scale virtual screening \cite{doi:10.1021/acs.jcim.0c00900}.

\subsection{Challenges in Biological Data Representations}

Small molecules/ligands in drug discovery interact with their protein targets in a variety of manners \cite{Hughes2011PrinciplesOE}. The most common interaction is the ligand burying itself within the core of the protein's ligand binding pocket, where the protein's native substrate usually resides \cite{CARPENTER20241320}. A ligand residing in the protein's ligand binding pocket either accelerates or decelerates the protein’s standard function, thereby eliciting the desired biological phenotype. Alternatively, ligands which reside on the surface of proteins, sometimes referred to as allosteric interactions, disrupt protein-protein interactions (PPI) and impact the usual function of a protein \cite{CARPENTER20241320}. 

Meaningful data representations of ligand binding pocket and allosteric binding interactions of small molecules with proteins remain a challenge for machine learning practitioners (Figure 3). Namely, the 3-dimensional topology and plasticity of ligand-protein interactions can be difficult to capture with traditional machine learning data structures such as lists, hash tables, and graphs. Further, it has been shown that a single atom change on a ligand can not only erode potency, but also lead to completely different mechanisms of action on the same protein \cite{doi:10.1021/ml500420y}.

\begin{figure}[!ht]
\begin{center}
\includegraphics[width=140mm]{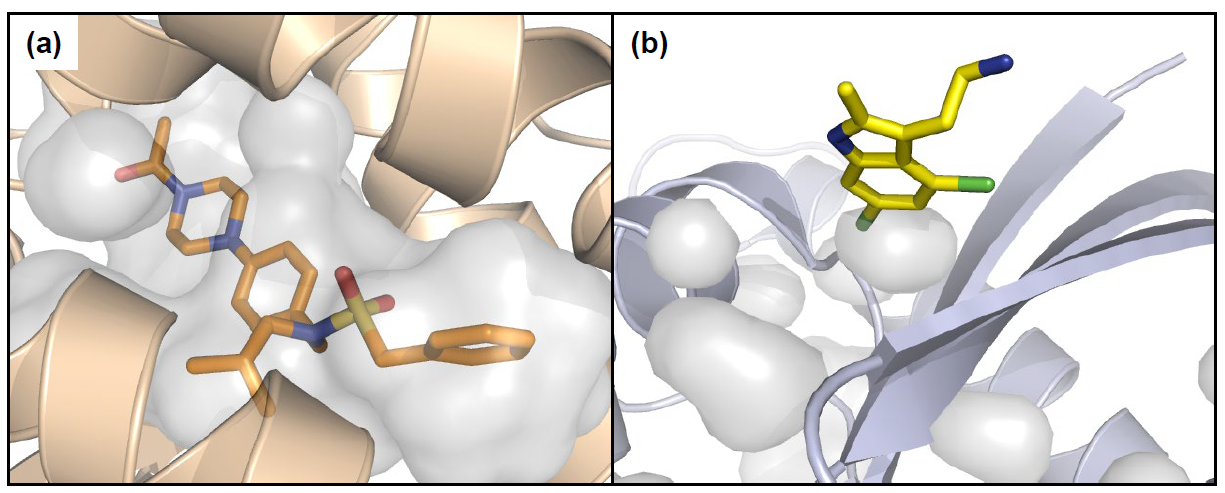}
\caption{Examples of (a) ligand binding pocket and (b) allosteric ligand-protein interactions. (a) Cocrystal structure (1.99 \AA) of a tertiary sulfonamide ligand (orange) in complex with human RORc-LBD (beige) [PDB: 4WQP]. (b) Cocrystal structure (2.39 \AA) of the small molecule ligand DCAI (yellow) in complex with human KRas4B-G12D (light blue) [PDB: 4DST]. Both images depict the ligand binding pockets of the respective proteins as transparent surfaces (light grey), and protein side chains are omitted for clarity. Notably, (a) exemplifies a deep ligand binding pocket within the protein, whereas (b) illustrates an allosteric interaction of the DCAI ligand on the protein surface. Further, (b) clearly lacks any significant binding pocket interactions between the ligand and protein, and the DCAI ligand disrupts the protein-protein interaction between the KRas and SOS proteins (SOS protein not shown).}
\label{pocketandallosteric}
\end{center}
\vskip -0.1in
\end{figure}

Modeling complicated multivariate phenomena, such as language or human behavior, via elegant, closed form equations can be challenging. Instead, it has been shown that the solutions to these complex problems resides in the “unreasonable effectiveness of data” \cite{Halevy2009TheUE}. 

Web-scale data has been the primary driver advancing deep learning methods \cite{Hoffmann2022TrainingCL, Kaplan2020ScalingLF}. These advances have resulted in impressive generalist computer vision models \cite{Voulodimos2018DeepLF}, image generation models \cite{Zhang2023TexttoimageDM}, and large language models which can generate human-like text \cite{Achiam2023GPT4TR, Anil2023PaLM2T}. Herein, we demonstrate that ligand-protein interaction affinities can be accurately predicted by coupling advances in instruction fine-tuned foundational pretrained generative small language models (SLMs) \cite{Fauber2024PretrainedGL} with the effectiveness of large-scale ligand-protein interaction affinity data.

\section{Methods}

\subsection{Public Data Sets}

Several publicly available data sets describe binary ligand-protein interactions where ligands are either “binders” represented as a $1$, or “non-binders” represented as a $0$, of a target protein. Examples of binary data sets include BioSNAP \cite{biosnapnets}, DrugBank \cite{Wishart2007DrugBankAK}, and the Yamanishi data set \cite{Yamanishi2008PredictionOD}.\footnote{http://web.kuicr.kyoto-u.ac.jp/supp/yoshi/drugtarget/ (accessed 28May2024).} 

Conversely, the Davis data set describes a range of ligand affinities for proteins with their corresponding pIC\textsubscript{50} affinity values \cite{Davis2011ComprehensiveAO}. The Davis data set contains 6,557 examples with 42 ligands and 272 protein kinases. The Davis data set has been utilized in LPI prior art, yet this small kinase-focused data set offers limited opportunity for generalization. 

A recent release of BindingDB (April 2024 version) contains 2M unique ligand-protein interactions and their corresponding K\textsubscript{d}, K\textsubscript{i}, IC\textsubscript{50}, and/or EC\textsubscript{50} affinity value(s) \cite{Gilson2015BindingDBI2}.\footnote{https://www.bindingdb.org/ (accessed 28April2024)} Specifically, this data set contains 2,020,737 unique examples with 1,203,453 ligands, and 6,480 proteins. We refer to this data set as BindingDB-2M.

\subsection{Our Data Sets}

We created an additional 1.5M examples of protein-ligand interactions and their corresponding affinity values to further expand on the Davis and BindingDB data sets. Our expanded data set was created from all entries in the United States National Institutes of Health (NIH) PubChem database \cite{Kim2015PubChemSA}. Our data set curation process is described in the Appendix section. 

We chose to gather additional data from PubChem as a majority of the entries in the BindingDB data set originated from the ChEMBL database \cite{10.1093/nar/gkad1004}, and only 4\% originated from the PubChem database (Figure 4). This difference in data sources offered an opportunity to complement the existing data within BindingDB with additional data from PubChem.

\begin{figure}[!ht]
\begin{center}
\includegraphics[width=100mm]{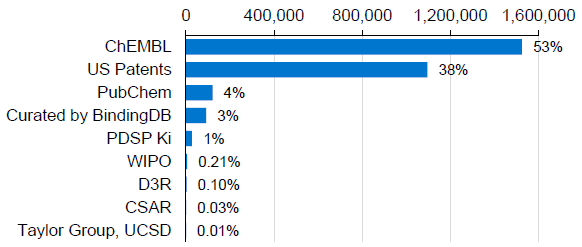}
\caption{Sources of the BindingDB ligand-protein interaction data set as of April 2024. Raw count values are shown on the x-axis, and the corresponding percentage of the total count for each data source are noted as labels on each bar of the plot.}
\label{bindingdbsources}
\end{center}
\vskip -0.1in
\end{figure}

Our mining of PubChem to create a new ligand-protein interaction affinity data set resulted in 1,478,702 unique examples with 927,688 ligands and 4,771 proteins. We refer to our new data set as LPI-1.5M (Table 1). Our LPI-1.5M data set was also merged with the BindingDB and Davis data sets, then all duplicate entries were removed, resulting in a final data set of 3,503,932 examples with 2,130,550 ligands and 6,732 proteins. We refer to our larger data set as LPI-3.5M. The LPI-1.5M and LPI-3.5M data sets both contained the ligand SMILES string, UNIPROT ID of the protein \cite{10.1093/nar/gkac1052}, amino acid sequence of the protein, and pIC\textsubscript{50} affinity value of the each ligand-protein interaction. 

\begin{table*}[!ht]
\vskip 0.1in
\begin{center}
\begin{small}
\begin{tabular}{lcccc}
\toprule
\specialcellcenter{Ligand-Protein Interaction \\ Affinity data set} & \specialcellcenter{Source} &
\specialcellcenter{Unique \\ Examples} & \specialcellcenter{Unique \\ Ligands} & \specialcellcenter{Unique \\ Proteins} \\
\midrule
Davis & Public & 6,557 & 42 & 272 \\
BindingDB-2M & Public & 2,020,737 & 1,203,453 & 6,480 \\
LPI-1.5M & Ours & 1,478,702 & 927,688 & 4,771 \\
LPI-3.5M & Ours & 3,503,932 & 2,130,550 & 6,732 \\
\bottomrule
\end{tabular}
\end{small}
\caption{Profiles of publicly available and our ligand-protein interaction affinity data sets. The values shown for BindingDB are as of April 2024.}
\end{center}
\vskip -0.2in
\end{table*}

\subsection{Data Formatting}

All pIC\textsubscript{50} values, regardless of the data set, were binned into five discrete ordinal affinity values corresponding a letter of the alphabet: A through E (Figure 5). The ordinal values included: A (pIC\textsubscript{50}$\ge 8$), B ($8 >$ pIC\textsubscript{50} $\ge 7$), C ($7 >$ pIC\textsubscript{50} $\ge 6$), D ($6 >$ pIC\textsubscript{50} $\ge 5$), and E ($ 5 >$ pIC\textsubscript{50}). Our machine learning studies used the alphabetical ordinal values, while instruction fine-tuning of pretrained generative small language models (SLMs) utilized these same alphabetical values and assigned them onomatopoeia consistent with the language of Dr. Seuss \cite{1970mrbrown}.

\subsection{Data Sampling}

Following best practices in machine learning, we randomly divided parent data sets into training/fine-tuning data and test data by sampling without replacement. The training/fine-tuning and testing data sets mirrored their parent data set ordinal affinity value distributions (Figure 5), only differing by $\pm 2 \%$ at most.

We varied the number of available fine-tuning data instances from 10,000 to 3.5M examples of ligand-protein interactions and their corresponding ordinal affinity values, by random selection without replacement from the parent pool of fine-tuning data instances for each instance cohort. The fine-tuning data instances were used for language model instruction fine-tuning. The language models were never exposed to the test data (\emph{i.e.}, out-of-sample "hold-out" data) during the fine-tuning process to avoid train/test data contamination. 

\begin{figure}[!ht]
\vskip 0.1in
\begin{center}
\includegraphics[width=130mm]{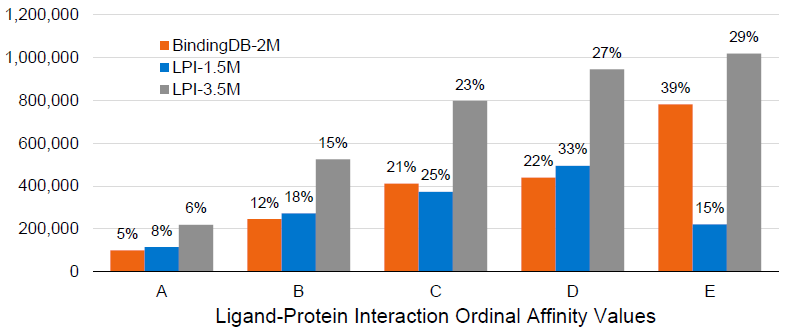}
\caption{Ordinal affinity value distributions of the BindingDB-2M (orange), LPI-1.5M (blue), and LPI-3.5M (grey) data sets. The ligand-protein interaction ordinal affinity values shown on the x-axis are: A (pIC\textsubscript{50}$\ge 8$), B ($8 >$ pIC\textsubscript{50} $\ge 7$), C ($7 >$ pIC\textsubscript{50} $\ge 6$), D ($6 >$ pIC\textsubscript{50} $\ge 5$), and E ($ 5 >$ pIC\textsubscript{50}). Raw count values are shown on the y-axis, and the corresponding percentage of the total data set for each class are noted as labels on each bar of the plot.}
\label{datasetdist}
\end{center}
\vskip -0.1in
\end{figure}

All data was formatted into an instruction-based format where the “instruction” was the input, and the “output” was the desired outcome. For example the instruction was, “Predict the potency of the following SMILES and UNIPROT sequences: N[C@H]1C[C@H]1c1ccc(NC(=O)c2ccccc2)cc1 and MENQEKASIAGHMFDVVVIGGGISGLSAAKLLTEYGVSVLVLEARDRVGGRTYTIRNEHVDYVD...,” and the corresponding output was, “choochoo”. The instruction formatting was consistent throughout the fine-tuning and testing data sets.

\subsection{Pretrained Foundational Small Language Models}

We selected the OPT (open pretrained transformer) family of pretrained foundational generative language models as the starting point for our studies \cite{Zhang2022OPTOP}. We also explored the GPT-Neo \cite{Black2021GPTNeoLS} and TinyStories \cite{Eldan2023TinyStoriesHS} families of language models. Both OPT-125m and GPT-Neo-125m contained 125M model parameters, whereas TinyStories-28M contained 28M model parameters. All models provided up to 2,048 positional embeddings for their inputs, permitting context for long SMILES strings and/or amino acid sequences which can be present in our method.

In our work, we defined model fine-tuning as initialization of a pretrained foundational language model followed by updates to the model weights and biases. In our fine-tuning setting, all language model parameters could undergo gradient updates -- there were no frozen layers nor adapters. In our prior work \cite{Fauber2024PretrainedGL}, we found the full fine-tuning approach was superior to adapter-based methods like LoRA (Low-Rank Adaptation) \cite{Hu2021LoRALA}. Other research groups have since confirmed our initial findings \cite{biderman2024lora}.

The prompt for the language models was consistent throughout our evaluation and across all models. The language model prompt was general and agnostic to the data set instructions. The prompt used for our evaluation was:
“Below is an instruction that describes a task. Write a response that appropriately completes the request. \#\#\# Instruction: \{instruction\} \#\#\# Response:”.

\subsection{Evaluation of Our Method}

We evaluated the performance of our fine-tuned language models on their ability to correctly provide the ordinal value prediction that exactly matched the ground truth ordinal affinity value in our test data set in a zero-shot setting. We also evaluated the ability of our fine-tuned models to correctly predict the exact ordinal value or $\pm 1$ ordinal value relative to the ground truth value (\emph{e.g.}, prediction of B when the ground truth was C). Flexibility in allowing the "near match" affinity value is consistent with the usage and performance of the FEP+ method \cite{schrodingerfeplivedesign2023, Ross2023TheMA}. It is also practical for the rank ordering of ligands in virtual screening campaigns.

We used our instruction fine-tuning and text generation framework for consistency in outcomes and scoring \cite{Fauber2024PretrainedGL}. There were no detectable deviations in our study when replicate training sessions and fine-tuned SLM text generation results were evaluated.  

\section{Results}

\subsection{ML Model Performance}

We explored the performance of statistical machine learning (ML) models on our LPI affinity prediction task. A training set of 100,000 LPI examples, and their corresponding ordinal affinity values, were drawn from the LPI-1.5M data set. 

The ligand SMILES strings were converted into both MACCS (Molecular ACCess System) fingerprint sparse embeddings \cite{doi:10.1021/ci010132r} and extended-connectivity "circular" fingerprint (ECFP) sparse embeddings \cite{doi:10.1021/ci100050t}. The protein amino acid sequences were converted into dense embeddings with the ESM2-3B (Evolutionary Scale Modeling 2) model \cite{doi:10.1126/science.ade2574}. These ligand and protein embedding techniques were selected due to their prevalence and performance in LPI binary affinity classification prior art \cite{Kimber2021DeepLI}. The ligand and protein embeddings were concatenated, then $\ell_2$-normalized. The same process was applied to a 10,000-example test set from the LPI-1.5M data set. The train and test data sets were unique with no overlap.

A support vector machines (SVM) machine learning model was selected for this analysis given its strong performance on imbalanced data sets \cite{Chakrabarti2022RobustHC}, which are often present in multinomial classification tasks such as ours (Figure 5).\footnote{https://scikit-learn.org/stable/modules/generated/sklearn.svm.LinearSVC (accessed 11June2024)} A one-versus-rest (OvR) instance of a linear kernel SVM was employed, thus enabling our multinomial classification task.\footnote{https://scikit-learn.org/stable/modules/generated/sklearn.multiclass.OneVsRestClassifier.html (accessed 11June2024)} Additional details for our data embedding and ML methods are described in the Appendix.

\begin{table*}[!ht]
\vskip 0.1in
\begin{center}
\begin{small}
\begin{tabular}{lccccc}
\toprule
\specialcellcenter{Machine Learning \\ Model} & 
\specialcellcenter{Ligand \\ Embedding \\ Model} & 
\specialcellcenter{Protein \\ Embedding \\ Model}  & \specialcellcenter{Dimension of \\ Ligand + Protein \\ Embedding}  & \specialcellcenter{\% Accuracy} & \specialcellcenter{\% Exact \\ Matches} \\
\midrule
OvR(LinearSVM) & ECFP & ESM2-3B & 4,608 & 7\% & 7\% \\
OvR(LinearSVM) & MACCS & ESM2-3B & 2,727 & 7\% & 7\%\\
\bottomrule
\end{tabular}
\end{small}
\caption{Performance of ML models in the conversion of 10,000 test instances of ligand embeddings and protein amino acid sequence embeddings into their corresponding predicted LPI ordinal affinity values from the LPI-1.5M data set. The ML model outputs were compared to their ground truth values for scoring.}
\end{center}
\vskip -0.1in
\end{table*}

The OvR instances of linear SVM models demonstrated 7\% overall accuracy and 7\% overall exact matches on our multinomial classification task for both ligand embedding techniques (Table 2). Additionally, both model instances produced 0\% exact matches for the A and B ordinal affinity values, and 1\%, 15\%, and 9\% exact matches for the ordinal affinity values C, D, and E, respectively. These results resemble the distribution of the parent LPI-1.5M data (Figure 5), yet lack sufficient utility in prioritizing ligands for progression in a drug discovery campaign.

\subsection{Language Model Baseline Performance}

We initially established a baseline for the pretrained foundational small language models on our LPI affinity prediction task. All models were incapable of performing our task with any detectable proficiency (Table 2). 

\begin{table*}[!ht]
\vskip 0.1in
\begin{center}
\begin{small}
\begin{tabular}{lccc}
\toprule
\specialcellcenter{Pretrained Foundational \\ Language Model} & \specialcellcenter{Language Model \\ Parameter Count} & \specialcellcenter{\% Accuracy} & \specialcellcenter{\% Exact \\ Matches} \\
\midrule
roneneldan/TinyStories-28M & 28M & 0\% & 0\% \\
facebook/opt-125m & 125M & 0\% & 0\% \\
EleutherAI/gpt-neo-125m & 125M & 0\% & 0\% \\
\bottomrule
\end{tabular}
\end{small}
\caption{Baseline performance of pretrained foundational small language models in the conversion of 10,000 test instances of ligand SMILES strings and protein amino acid sequences into their corresponding predicted LPI ordinal affinity values from the LPI-1.5M data set. The model outputs were compared to their ground truth values for scoring. The language models are described by their \texttt{HuggingFace.co} repo names (accessed 30May2024).}
\end{center}
\vskip -0.1in
\end{table*}

We recognize that fine-tuning language models over multiple epochs may obliterate some portion of information that resides within the pretrained foundational language model. This potential change did not concern us as our objective was to create specialized language models from pretrained foundational language models, with the objective of effectively executing a highly specialized task that the original pretrained foundational models were incapable of performing. 

\subsection{Performance of Our Method}

The OPT-125M pretrained small language model was instruction fine-tuned on 100,000 training examples drawn from the LPI-1.5M data set. We observed a significant improvement in the performance of our fine-tuned SLM on our LPI affinity prediction task versus the baseline model on a test set of 10,000 examples from the LPI-1.5M data set. Our fine-tuned SLM achieved 37\% overall accuracy and 37\% overall exact matches on our task. Notably, our fine-tuned SLM achieved 14\%, 36\%, 64\%, and 22\% exact matches for the ordinal affinity values B, C, D, and E, respectively (Figure 6). These results were significantly better than the ML results (Table 2) and baseline language model results (Table 3) on the same train/test data sets. 

\begin{figure}[!ht]
\vskip 0.1in
\begin{center}
\includegraphics[width=110mm]{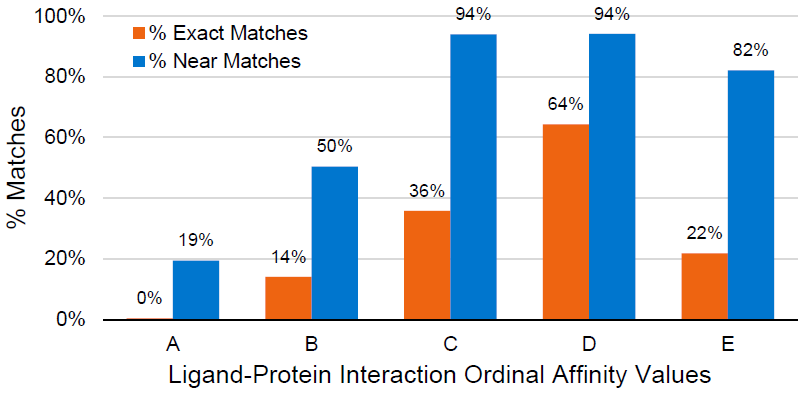}
\caption{Performance of our instruction fine-tuned OPT-125M SLM on our LPI affinity prediction task with 100,000 training examples from the LPI-1.5M data set. The fine-tuned model performance was assessed with a 10,000-example test set drawn from the LPI-1.5M data set. The model outputs were compared to their ground truth values for scoring as either: 1) a \% exact match (orange) or 2) a \% near match (blue). A near match was defined as an equal ordinal affinity value or $\pm 1$ value relative to the ground truth. The ordinal affinity values shown on the x-axis are: A (pIC\textsubscript{50}$\ge 8$), B ($8 >$ pIC\textsubscript{50} $\ge 7$), C ($7 >$ pIC\textsubscript{50} $\ge 6$), D ($6 >$ pIC\textsubscript{50} $\ge 5$), and E ($ 5 >$ pIC\textsubscript{50}).}
\label{ourmethod100k}
\end{center}
\vskip -0.1in
\end{figure}

Relaxing the scoring criteria to a predicted ordinal affinity value equal to or $\pm 1$ value relative to the ground truth, as is regularly employed in the FEP+ method \cite{schrodingerfeplivedesign2023, Ross2023TheMA}, resulted in impressive outcomes with our method. With the relaxed "near match" criteria, we achieved an 77\% overall accuracy and all ordinal affinity values achieved 19-94\% near matches relative the the ground truth with our method (Figure 6). The relaxed criteria of a near match is reasonable for the prioritization of ligands in virtual screening, and is likely why this practice was introduced by FEP+ practitioners.
  
\subsection{Influence of Data Set Size on Our Method}

Increasing the number of training examples during instruction fine-tuning of pretrained foundational small language models resulted in monotonically increasing performance, as assessed by overall \% accuracy and overall \% exact matches (Table 4). The OPT models were of comparable performance to the GPT-Neo models on our LPI affinity prediction task. The performance of the TinyStories models were below those of the OPT and GPT-Neo models, yet it was notable that these 28M parameter fine-tuned SLMs were able to complete our LPI affinity prediction task with a reasonable level of proficiency. 

\begin{table*}[!ht]
\vskip 0.1in
\begin{center}
\begin{small}
\begin{tabular}{lcrcc}
\toprule
\specialcellcenter{Pretrained Foundational \\ Language Model} &
\specialcellcenter{Language Model \\ Parameters} & \specialcellcenter{Instruction Fine-Tuning \\ Data Set Size} & \specialcellcenter{\% Accuracy} & \specialcellcenter{\% Exact \\ Matches} \\
\midrule
roneneldan/TinyStories-28M & 28M & 10,000 examples & 31\% & 31\% \\
EleutherAI/gpt-neo-125m & 125M & 10,000 examples & 33\% & 33\% \\
facebook/opt-125m & 125M & 10,000 examples & 34\% & 34\% \\
roneneldan/TinyStories-28M & 28M & 100,000 examples & 32\% & 32\% \\
EleutherAI/gpt-neo-125m & 125M & 100,000 examples & 36\% & 36\% \\
facebook/opt-125m & 125M & 100,000 examples & 37\% & 37\% \\
roneneldan/TinyStories-28M & 28M & 1,000,000 examples & 35\% & 35\% \\
EleutherAI/gpt-neo-125m & 125M & 1,000,000 examples & 35\% & 35\% \\
facebook/opt-125m & 125M & 1,000,000 examples & 38\% & 38\% \\
\bottomrule
\end{tabular}
\end{small}
\caption{Influence of increasing instruction fine-tuning examples. Pretrained foundational language models were instruction fine-tuned on increasing numbers of training examples from the LPI-1.5M data set. The fine-tuned models were assessed in their conversion of 10,000 test instances of ligand SMILES strings and protein amino acid sequences into their corresponding predicted LPI ordinal affinity values from the LPI-1.5M data set. The foundational language models are described by their \texttt{HuggingFace.co} repo names (accessed 30May2024).}
\end{center}
\vskip -0.1in
\end{table*}

As we increased the number of instruction fine-tuning examples, not only did the overall accuracy of our LPI affinity predictions consistently improve across all fine-tuned models (Table 4), but their accuracy in predicting specific LPI ordinal affinity values also saw a corresponding improvement. Specifically, the instruction fine-tuning of the OPT-125M pretrained foundational language model on increasing quantities of examples from the LPI-1.5M data set resulted in improved LPI affinity predictions, as measured by F1 scores, across all LPI ordinal affinity values (Figure 7). 

The F1 score is a harmonic mean of the per-class precision (the ratio of true positives to false positives) and the per-class recall (\% exact matches). The F1 score was useful in observing the improvements in LPI affinity predictions for each class as the number of fine-tuning examples increased. As the number of fine-tuning examples increased, the per-class distribution of the F1 score began to mirror that of the parent LPI-1.5M data set distribution (Figure 5). This transformation illustrated that the models were learning how to correctly predict all LPI ordinal affinity values as the number of fine-tuning examples increased.

\begin{figure}[!ht]
\vskip 0.1in
\begin{center}
\includegraphics[width=130mm]{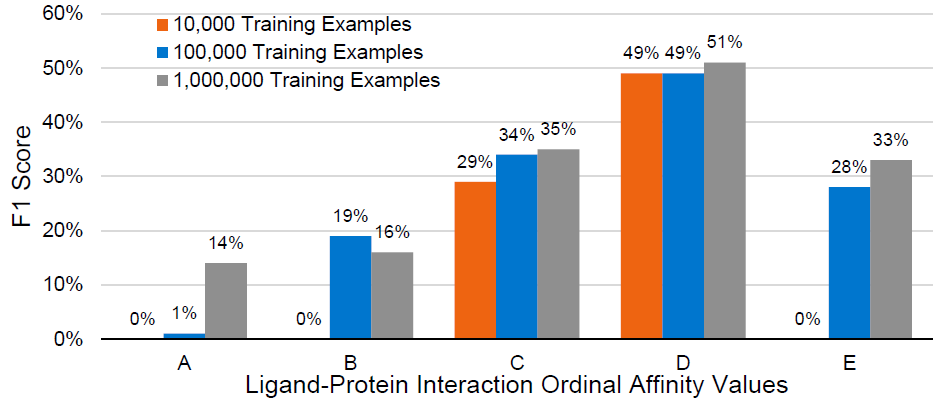}
\caption{Scaling of instruction fine-tuning data and effects on LPI affinity predictions. A pretrained OPT-125M language model was instruction fine-tuned on our LPI affinity prediction task with either 10,000 (orange), 100,000 (blue), or 1,000,000 (grey) training examples from the LPI-1.5M data set. The fine-tuned model performance was assessed with a 10,000-example test set drawn from the LPI-1.5M data set. The model outputs were compared to their ground truth for scoring. The ordinal affinity values shown on the x-axis are: A (pIC\textsubscript{50}$\ge 8$), B ($8 >$ pIC\textsubscript{50} $\ge 7$), C ($7 >$ pIC\textsubscript{50} $\ge 6$), D ($6 >$ pIC\textsubscript{50} $\ge 5$), and E ($ 5 >$ pIC\textsubscript{50}).}
\label{scalingbyclass}
\end{center}
\vskip -0.1in
\end{figure}

Further scaling our instruction fine-tuning training data set to 3.5M examples drawn from the LPI-3.5M data set resulted in a higher accuracy model than the LPI-1.5M-trained model shown in Figure 6. Our 125M parameter instruction fine-tuned SLM demonstrated 44\% overall accuracy and 44\% overall exact matches on 10,000 test examples drawn from the LPI-3.5M data set. Our fine-tuned SLM achieved 19\%, 7\%, 39\%, 49\%, and 74\% exact matches for the ordinal affinity values A, B, C, D, and E, respectively (Figure 8). Relaxing the scoring criteria to a "near match" predicted ordinal affinity value equal to or $\pm 1$ value relative to the ground truth, resulted in a 79\% overall accuracy and all ordinal affinity values achieving 28-97\% near matches relative the the ground truth (Figure 8). 

Our results were noteworthy as a recent retrospective analysis of $\sim 13,000$ FEP+ calculations demonstrated that under the same "near match" criteria, they achieved a 58\% overall accuracy relative to ground truth values \cite{Ross2023TheMA}. Our substantial improvements in accurately predicting LPI affinities, combined with the simplicity and high throughput of our method, represents an appreciable advancement in accurately predicting a range of ligand-protein interaction affinities.

\begin{figure}[!ht]
\vskip 0.1in
\begin{center}
\includegraphics[width=110mm]{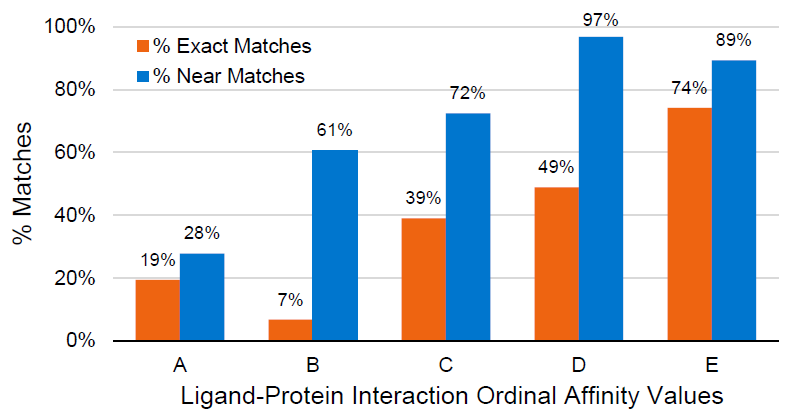}
\caption{Performance of our instruction fine-tuned OPT-125M SLM on our LPI affinity prediction task with 3.5M training examples from the LPI-3.5M data set. The model performance was assessed with a 10,000-example test set drawn from the LPI-3.5M data set. The model outputs were compared to their ground truth values for scoring as either: 1) a \% exact match (orange) or 2) a \% near match (blue). A near match was defined as an equal ordinal affinity value or $\pm 1$ value relative to the ground truth. The ordinal affinity values shown on the x-axis are: A (pIC\textsubscript{50}$\ge 8$), B ($8 >$ pIC\textsubscript{50} $\ge 7$), C ($7 >$ pIC\textsubscript{50} $\ge 6$), D ($6 >$ pIC\textsubscript{50} $\ge 5$), and E ($ 5 >$ pIC\textsubscript{50}).}
\label{ourmethod3.5m}
\end{center}
\vskip -0.1in
\end{figure}

The results achieved by our fine-tuned SLM somewhat mirror the ordinal affinity value distribution of the LPI-3.5M parent data set (Figure 5). We note that the \% exact match value for ordinal affinity value B is lesser than what might be expected based on its cohorts (Figure 8). The rationale for this difference is unknown, but it is likely tied to an artifact in LPI affinity prediction results for the BindingDB-2M data set, which comprises a portion of the LPI-3.5M data set (\emph{see} Appendix). Overall, these results are better than the results from the SLM fine-tuned on 100,000 examples drawn from LPI-1.5M, demonstrating improved pan-class affinity prediction performance and highlighting the benefits of a larger training data set.

\subsection{Ablation Studies}

We conducted training data ablation studies to assess the importance of both the ligand SMILES string inputs and the protein amino acid sequence inputs in the instruction fine-tuning of pretrained small language models. We utilized 100,000 training examples from our LPI-1.5M data set for this ablation study. The training examples contained either: 1) ligand and protein inputs; 2) only ligand inputs (\emph{i.e.}, SMILES strings); or 3) only protein inputs (\emph{i.e.}, amino acid sequences). 

Three distinct models were created by instruction fine-tuning an OPT-125M pretrained language model on one of the three training data sets. The resulting three fine-tuned SLMs were then evaluated on our LPI affinity prediction task with 10,000 test examples drawn from the LPI-1.5M data set.

Our analysis revealed that only ligand inputs (34\% overall accuracy) and only protein inputs (32\% overall accuracy) were unable to achieve similar LPI affinity prediction performance to the ligand \emph{and} protein inputs (37\% overall accuracy). This led us to conclude that both the ligand and protein inputs were valuable in effectively predicting LPI affinities. 

The prediction results for the protein only inputs more closely mirrored the performance of the ligand and protein inputs when analyzed at the LPI ordinal affinity value level (Figure 9). These results demonstrated that the protein inputs might convey more influence in LPI affinity prediction outcomes than the ligand inputs. The observed dependency on the protein inputs was anticipated as the protein inputs occupied 83\% of the total prompt on average, whereas the ligand inputs only occupied 8\% of the total prompt on average, as measured by character count in the LPI-1.5M data set.\footnote{The remaining 9\% of the total prompt was attributed to the consistent instruction prompt for each input, as described in the Methods/Data Sampling section.} This observation was unique to our instruction fine-tuned language model setting, as others have observed a stronger outcome dependence on the ligand inputs for ML-based LPI binary (\emph{e.g.}, binder/non-binder) affinity predictions \cite{doi:10.1021/acs.jcim.3c01208}. 

\begin{figure}[!ht]
\vskip 0.1in
\begin{center}
\includegraphics[width=130mm]{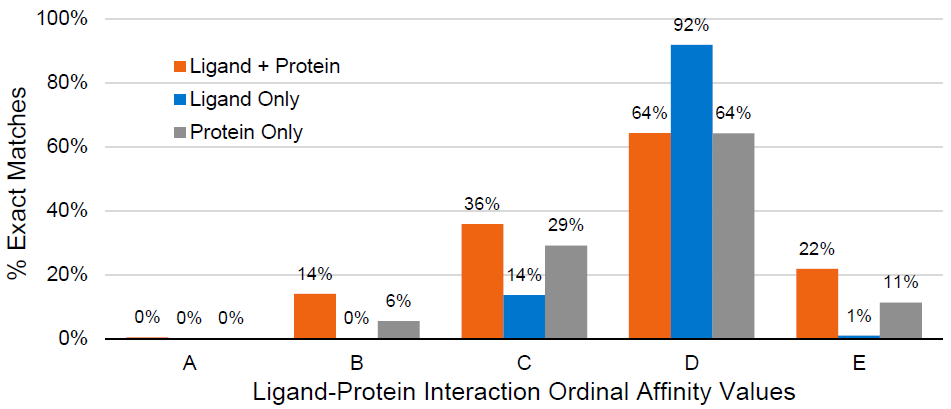}
\caption{Ablation studies for the instruction fine-tuning of language models with selected training data inputs. A pretrained OPT-125M language model was instruction fine-tuned on our LPI affinity prediction task with 100,000 training examples from the LPI-1.5M data set. The examples contained either: 1) ligand and protein inputs (orange); 2) only ligand inputs (blue); or 3) only protein inputs (grey). The model performance was assessed with a 10,000-example test set drawn from the LPI-1.5M data set. The model outputs were compared to their ground truth for scoring. The ordinal affinity values shown on the x-axis are: A (pIC\textsubscript{50}$\ge 8$), B ($8 >$ pIC\textsubscript{50} $\ge 7$), C ($7 >$ pIC\textsubscript{50} $\ge 6$), D ($6 >$ pIC\textsubscript{50} $\ge 5$), and E ($ 5 >$ pIC\textsubscript{50}).}
\label{ablations}
\end{center}
\vskip -0.1in
\end{figure}

\section{Discussion}

Our results demonstrated a clear improvement over ML and FEP+ based approaches in accurately predicting a range of ligand-protein interaction affinities. Our method also illustrated that powerful specialized models can be created with the instruction fine-tuning of pretrained foundational language models. Our method is practical and useful for the \emph{in silico} evaluation and prioritization of ligands for drug discovery campaigns whether a practitioner chooses to use the \% exact match or \% near match evaluation criteria. 

We note that the performance of our method on the A and B ligand-protein interaction ordinal affinity value predictions was below that of the other ordinal affinity values: C, D, and E (Figures 6 and 8). Yet, these results were anticipated as the ordinal affinity values A and B had low representation in the parent data sets (Figure 5). This low representation of molecules with potent LPI affinity values (\emph{e.g.}, pIC\textsubscript{50} $\ge 8$) relative to less potent analogs (\emph{e.g.}, pIC\textsubscript{50} $< 5$) likely mirrors the distribution of LPI affinity results found in biotechnology and pharmaceutical company databases, as inactive ligands or weak binders are far more common than potent binders of a target protein. Models learn from their training data, thus inclusion of additional potent LPI pairs might improve the prediction performance on the LPI ordinal affinity classes A and B.

We explored the performance of our method on test (\emph{i.e.}, out-of-sample or hold-out) data to avoid train/test data contamination. We view this as a reasonable evaluation framework for our and other LPI affinity prediction methods. 

Other research groups have proposed holding out distinct clusters of ligands and classes of proteins from the training/fine-tuning data sets. Thereby, allowing rigorous testing of LPI affinity prediction methods, ensuring that entire ligand clusters and protein classes are out-of-sample relative to the model training/fine-tuning data \cite{AtasGuvenilir2023HowTA, Park2012AFI}. Such an idealistic paradigm would indeed be useful in evaluating a model's ability to extrapolate into entirely new classes of previously unseen/untested ligands and proteins. Yet, this proposed approach is unrealistic in our current data environment.

The assay results from the \emph{in vitro} evaluation of ligand-protein interactions, and their corresponding affinities, are not available at extremely large scale (\emph{e.g.}, internet scale). Additionally, the results of these assays are sometimes published in journal articles or patents, but most results remain as closely guarded intellectual property of pharmaceutical and biotechnology companies. The shortage of 100M-scale to 1B-scale LPI affinity data sets will greatly hamper the ability to create powerful generalist models which can effectively extrapolate to accurately predict affinities of previously unseen/untested ligands, proteins, and their interactions. Rather, it is reasonable to continue to assess LPI affinity prediction methods using traditional machine learning best practices of train/test data sets until much larger LPI data sets become publicly available.  

\section{Conclusion}

We have demonstrated that instruction fine-tuned pretrained language models can accurately predict a range of ligand-protein affinities. Our results further demonstrated that pretrained foundational language models, and their architectures, can serve as general learning frameworks for a novel task of which the base model was incapable of performing. 

Our results illustrated a clear improvement over ML and FEP+ based approaches in accurately predicting a range of ligand-protein interaction affinities. Our method is practical and useful for the \emph{in silico} evaluation and prioritization of ligands for drug discovery campaigns. Our method can prove valuable whether a practitioner chooses to use the \% exact match or \% near match evaluation criteria. Additionally, our method is simple to implement as it only requires the SMILES string of the ligand and amino acid sequence of the target protein. We demonstrated that our approach can be generalized across many different open-source pretrained foundational language models. 

Specifically, we demonstrated that instruction fine-tuning pretrained SLMs with 10,000 to 3.5M examples resulted in the accurate prediction of a range of ligand-protein interaction affinities. Increasing the instruction fine-tuning examples can impart additional performance improvements in predicting LPI interaction affinities. It is likely that the prediction performance of language model based LPI affinity prediction methods like ours will continue to scale as instruction fine-tuning data sets grow larger.

\section{Acknowledgements}

The author would like to thank Anas Bricha and Neil Cameron for supporting this project. The author would also like to thank Deepayan Chakrabarti, Hans Purkey, and Dan Sutherlin for their insights, and Guy Laporte for providing access to the computational infrastructure to conduct these studies. The author declares no financial interests nor conflicts. 

\newpage

\bibliography{bib-llms}
\bibliographystyle{icml2020}

\newpage

\appendix

\setcounter{table}{0}
\setcounter{figure}{0}
\renewcommand{\thetable}{A\arabic{table}}
\renewcommand{\thefigure}{A\arabic{figure}}

\section{Appendix}
\subsection{Computational Infrastructure and Code}

The results described in this article were carried out using a Dell Technologies PowerEdge C4140 server with 4 x V100 NVIDIA\textsuperscript{\textregistered} SXM GPU cards with 32 GB VRAM each and NVLink\textsuperscript{TM} connectivity. There were 2 x Intel\textsuperscript{\textregistered} Xeon\textsuperscript{\textregistered} processors on the server with 1.5 TB of CPU RAM. 

The server was configured with the Ubuntu v22.04 Linux operating system, Anaconda v23.1.0, NVIDIA\textsuperscript{\textregistered} CUDA v12.2, and NVIDIA\textsuperscript{\textregistered} drivers v535.54.03. Additional python dependencies included: \texttt{accelerate v0.25.0},
\texttt{biopython v1.83},
\texttt{deepchem v2.7.1},
\texttt{scikit-learn v1.3.0}, 
\texttt{rdkit v2023.3.3},
\texttt{torch v2.1.1}, and \texttt{transformers v4.36.2}.

The Stanford ALPACA language model code was git cloned directly from \url{https://github.com/tatsu-lab/stanford_alpaca} (accessed 30Dec2023). The \texttt{train.py} file in the GitHub repo, along with our corresponding instruction fine-tuning data set, was used to instruction fine-tune the language models in our study. The language model fine-tuning code was executed via the command line interface (CLI).

As an example, the following CLI command was used to instruction fine-tune a pretrained foundational language model on 4 GPUs:

\begin{lstlisting}
torchrun --nproc_per_node=4 [TRAINING_PY_FILE] \
    --model_name_or_path [HUGGINGFACE_MODEL_NAME] \
    --data_path [DATA_PATH_TO_FORMATTED_JSON_FILE] \
    --bf16 False \
    --output_dir [OUTPUT_DIRECTORY] \
    --overwrite_output_dir True \
    --num_train_epochs 3 \
    --per_device_train_batch_size 4 \
    --per_device_eval_batch_size 4 \
    --gradient_accumulation_steps 8 \
    --save_strategy "steps" \
    --save_steps 5000 \
    --save_total_limit 1 \
    --learning_rate 2e-4 \
    --weight_decay 0. \
    --warmup_ratio 0.03 \
    --lr_scheduler_type "cosine" \
    --seed 41 \
    --logging_steps 1 \
    --tf32 False
\end{lstlisting}

\newpage

\subsection{Data Set Curation for LPI-1.5M and LPI-3.5M}

We created a data set of 1.5M examples of protein-ligand interactions and their corresponding affinity values to further expand on the Davis \cite{Davis2011ComprehensiveAO} and BindingDB \cite{Gilson2015BindingDBI2} LPI data sets. Our expanded LPI data set was created from all entries in the United States National Institutes of Health (NIH) PubChem database as of 08Feb2024 \cite{Kim2015PubChemSA}.\footnote{https://pubchem.ncbi.nlm.nih.gov/ (accessed 08Feb2024)} 

All available assay data was collected from the PubChem site for all compound identification values (CIDs), then filtered to those entries which contained either an IC\textsubscript{50}, EC\textsubscript{50}, AC\textsubscript{50}, K\textsubscript{i}, or K\textsubscript{d} value.\footnote{https://ftp.ncbi.nlm.nih.gov/pubchem/Bioassay/CSV/Data/ (Accessed on 08Feb2024)} If an assay did not demonstrate a range of affinity values for different compounds (\emph{e.g.}, all compounds were inactive), the assay was omitted from the data set. If a CID contained multiple affinity values for the same assay, the mean affinity value was carried forward. 

The PubChem assay results for LPI affinities were not in uniform units, as some were reported in molar (M) concentrations, while others were reported in millimolar (mM) concentrations. If we detected the range of LPI affinities were outside of the nanomolar (nM) to micromolar (\textmu M) range of affinities for a single assay, then those affinity results were normalized to the nM-to-\textmu M range of affinities for that individual assay.

The amino acid sequences associated with the PubChem assay results were mined from the NIH Entrez Molecular Sequence Database System using the UNIPROT ID of the assay target protein as the retrieval key \cite{10.1093/nar/gkac1052}.\footnote{http://www.ncbi.nlm.nih.gov/Entrez/ (accessed 13June2024)} Our mining of PubChem to create a new ligand-protein interaction affinity data set resulted in 1,478,702 unique examples with 927,688 ligands and 4,771 proteins. We refer to this data set as LPI-1.5M. In this data set, the average length of ligand SMILES strings was 68 characters, and the average length of protein amino acid sequences was 667 characters.

Our LPI-1.5M data set was also merged with the BindingDB (April 2024 version) and Davis data sets, then all duplicate entries were removed, resulting in a final data set of 3,503,932 examples with 2,130,550 ligands and 6,732 proteins. We refer to this larger data set as LPI-3.5M. The LPI-1.5M and LPI-3.5M data sets both contained the ligand SMILES string, UNIPROT ID of the protein, amino acid sequence of the protein, and pIC\textsubscript{50} affinity value of the each ligand-protein interaction. 

All pIC\textsubscript{50} values, regardless of the data set, were binned into five discrete ordinal values corresponding a letter of the alphabet: A through E (Figure 5). The ordinal values included: A (pIC\textsubscript{50}$\ge 8$), B ($8 >$ pIC\textsubscript{50} $\ge 7$), C ($7 >$ pIC\textsubscript{50} $\ge 6$), D ($6 >$ pIC\textsubscript{50} $\ge 5$), E ($ 5 >$ pIC\textsubscript{50}). Our machine learning studies used the alphabetical ordinal values, while the instruction fine-tuned small foundational pretrained generative language models (SLMs) utilized these same alphabetical values and assigned them onomatopoeia consistent with the language of Dr. Seuss \cite{1970mrbrown}.

The A through E onomatopoeia utilized for the instruction fine-tuning of the pretrained foundational small language models were A $\rightarrow$ "achoo," B $\rightarrow$ "blurpblurp," C $\rightarrow$ "choochoo," D $\rightarrow$ "dibbledopp," E $\rightarrow$ "eekeek." We chose to encode the target predictions as onomatopoeia given that generative language models predict the next most likely token in a sequence, and we wished for the target predictions to be semantically distinct from the inputs. We verified their semantic differences by computing and comparing the inner products of the mean-pooled penultimate OPT-125M model layer outputs for the various tokenized onomatopoeia.

\newpage

\subsection{Machine Learning Data Representations, Models, and Metrics}

We explored the performance of statistical machine learning (ML) models on our LPI affinity prediction task. A training set of 100,000 LPI examples, and their corresponding ordinal affinity values, were drawn from the LPI-1.5M data set. 

The ligand SMILES strings were converted into both MACCS (Molecular ACCess System) fingerprint sparse embeddings \cite{doi:10.1021/ci010132r} and extended-connectivity "circular" fingerprint (ECFP) sparse embeddings \cite{doi:10.1021/ci100050t}. The RDKit python toolkit with default settings was used to generate both embeddings.\footnote{https://www.rdkit.org/ (accessed 30April2024)} The 167-dimensional MACCS embeddings of the ligands were 67\% sparse on average, and the 2,048-dimensional ECFP embeddings of the ligands were 97\% sparse on average with this training data set.

The protein amino acid sequences were converted into dense embeddings with the ESM2-3B model \cite{doi:10.1126/science.ade2574}. The "esm2\_t36\_3B\_UR50D" instance of the ESM2-3B models was used with the default parameters.\footnote{https://huggingface.co/facebook/esm2\_t36\_3B\_UR50D (accessed 24May2024)} To generate dense embeddings of the protein amino acid sequences, the amino acid sequence was tokenized with the ESM2-3B model's tokenizer. The tokenzied sequence was then sent to the ESM2-3B model and the penultimate layer of the model output, but for the first and final columns of the output, was subjected to a column-wise mean-pooling operation to provide a 2,560-dimensional dense embedding of the initial protein amino acid sequence input.

The ligand and protein embeddings were concatenated along the same axis in that order, then $\ell_2$-normalized. The same process was applied to a 10,000-example test set from the LPI-1.5M data set. As noted earlier, the train and test data sets were unique with no overlap.

A support vector machines (SVM) machine learning model was selected for this analysis given its strong performance on imbalanced data sets \cite{Chakrabarti2022RobustHC}, which are often present in multinomial classification tasks such as ours (Figure 5).\footnote{https://scikit-learn.org/stable/modules/generated/sklearn.svm.LinearSVC (accessed 11June2024)} A one-versus-rest (OvR) instance of a linear kernel SVM was employed, thus enabling our multinomial classification task.\footnote{https://scikit-learn.org/stable/modules/generated/sklearn.multiclass.OneVsRestClassifier.html (accessed 11June2024)} Both the linear SVM and OvR instances were used with their default hyperparameters/settings.

Model performance was assessed using the python \texttt{scikit-learn} "accuracy\_score()" and "classification\_report()" modules.\footnote{https://scikit-learn.org/stable/api/sklearn.metrics.html (accessed 24May2024)} The percentage of correctly classified instances per class (\emph{i.e.}, \% exact matches for each ordinal affinity value) were scored via the per-class recall metric.

\newpage

\subsection{Language Model Text Generation Configuration}

The same language model fine-tuning and generation configurations were utilized throughout our studies, and only single-parameter changes were permitted, as annotated in the tables, when comparing methods. Language model text generation was conducted via the HuggingFace \texttt{transformers} library. Transformers \texttt{GenerationConfig()} was set to the default parameters, along with:
\begin{itemize}
    \item \texttt{num\_beams = 2}, 
    \item \texttt{repetition\_penalty = 1.3}, 
    \item \texttt{do\_sample = False} (for consistent output generation), 
    \item \texttt{early\_stopping = True}, 
    \item \texttt{max\_time = 10}, and 
    \item \texttt{length\_penalty = 0.4}.
\end{itemize}

In prior studies, we found the above configuration parameters provided stable and reproducible text generation \cite{Fauber2024PretrainedGL}. The text generation prompt and the general prompt used in the language model fine-tuning process were identical: 
\\
"\texttt{Below is an instruction that describes a task. Write a response that appropriately completes the request. \#\#\# Instruction: \{instruction\} \#\#\# Response: \{output\}}". 

Although not always necessary, we enforced truncation of the output text for all models to ensure consistency in outcomes. Truncation of the OPT model text output returned all text following the "\texttt{\#\#\# Response:}" string. Similarly, the GPT-Neo and TinyStories families of models truncated the output to the text following the "\texttt{<|endoftext|>}" string.

\newpage

\subsection{LPI Affinity Predictions for BindingDB-2M data set}

A pretrained OPT-125M language model was instruction fine-tuned on our LPI affinity prediction task with either 10,000, 100,000, or 1,000,000 training examples from the BindingDB-2M data set to create three distinct models. The prediction performance of the three fine-tuned models were assessed with a 10,000-example test set drawn from the BindingDB-2M data set. The model outputs were compared to their ground truth for scoring. 

We noted that increasing the number of training/fine-tuning examples increased the \% exact matches for most LPI ordinal affinity values (Figure A1). Yet, we also noted that the performance of the three different instruction fine-tuned SLMs did not mirror the distribution of the parent BindingDB-2M data set (Figure 5). Rather, the affinity prediction results for each LPI ordinal affinity value resulted in an overall bimodal distribution (Figure A1). The reasons for the observed bimodal distribution in LPI affinity predictions were unclear, but they were consistent outcomes with all three models that were fine-tuned on the BindingDB-2M data set. 

\begin{figure}[!ht]
\vskip 0.1in
\begin{center}
\includegraphics[width=130mm]{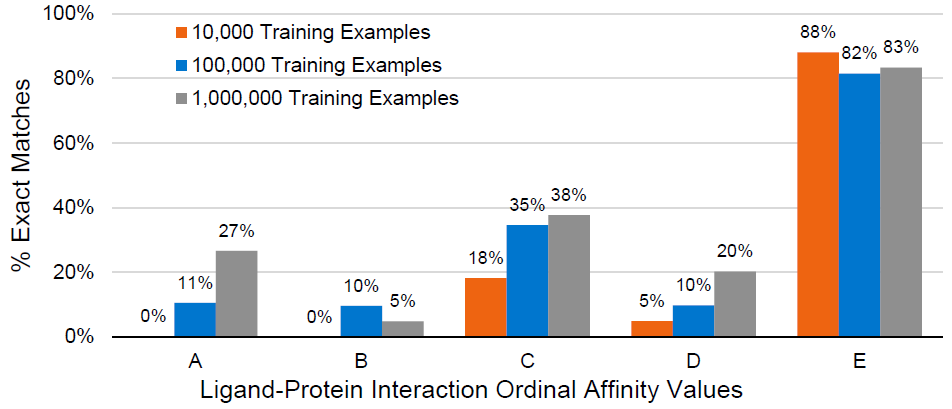}
\caption{LPI affinity predictions for three different SLMs fine-tuned/trained and tested on data from the BindingDB-2M data set. A pretrained OPT-125M language model was instruction fine-tuned on our LPI affinity prediction task with either 10,000 (orange), 100,000 (blue), or 1,000,000 (grey) training examples from the BindingDB-2M data set. The fine-tuned model performance was assessed with a 10,000-example test set drawn from the BindingDB-2M data set. The model outputs were compared to their ground truth for scoring. The ordinal affinity values shown on the x-axis are: A (pIC\textsubscript{50}$\ge 8$), B ($8 >$ pIC\textsubscript{50} $\ge 7$), C ($7 >$ pIC\textsubscript{50} $\ge 6$), D ($6 >$ pIC\textsubscript{50} $\ge 5$), and E ($ 5 >$ pIC\textsubscript{50}).}
\label{bindingdbpred}
\end{center}
\vskip -0.1in
\end{figure}

\end{document}